\documentclass[lettersize,journal]{IEEEtran}
\usepackage{amsmath,amsfonts}
\usepackage{algorithmic}
\usepackage{algorithm}
\usepackage{array}
\usepackage[caption=false,font=normalsize,labelfont=sf,textfont=sf]{subfig}
\usepackage{textcomp}
\usepackage{stfloats}
\usepackage{url}
\usepackage{verbatim}
\usepackage{graphicx}
\usepackage{cite}
\usepackage{amsmath}
\usepackage{amssymb}
\usepackage{mathtools}
\usepackage{amsthm}
\usepackage{colortbl}
\usepackage{xcolor}
\usepackage{hyperref}
\usepackage{multirow}
\begin{document}

\title{An Information Compensation Framework\\ for Zero-Shot Skeleton-based Action Recognition}


\author{\IEEEauthorblockN{Haojun Xu, Yan Gao, Jie Li, and Xinbo Gao,~\IEEEmembership{Fellow,~IEEE}}
	
\thanks{Haojun Xu, Yan Gao, and Jie Li are with the State Key Laboratory of Integrated Services Networks, School of Electronic Engineering, Xidian University, Xi'an 710071, China (e-mail: \{haojunxu, yangao\}@stu.xidian.edu.cn; leejie@mail.xidian.edu.cn).} 

\thanks{Xinbo Gao is with the State Key Laboratory of Integrated Services Networks, School of Electronic Engineering, Xidian University, Xi'an 710071, China (e-mail: xbgao@mail.xidian.edu.cn) and with the Chongqing Key Laboratory of Image Cognition, Chongqing University of Posts and Telecommunications, Chongqing 400065, China (e-mail: gaoxb@cqupt.edu.cn). (Corresponding author: Xinbo Gao, Jie Li.)}}

\markboth{Journal of IEEE Transactions on Multimedia}%
{Shell \MakeLowercase{\textit{et al.}}: A Sample Article Using IEEEtran.cls for IEEE Journals}


\maketitle

\begin{abstract}
Zero-shot human skeleton-based action recognition aims to construct a model that can recognize actions outside the categories seen during training. Previous research has focused on aligning sequences' visual and semantic spatial distributions. 
However, these methods extract semantic features simply. They ignore that proper prompt design for rich and fine-grained action cues can provide robust representation space clustering. In order to alleviate the problem of insufficient information available for skeleton sequences, we design an information compensation learning framework from an information-theoretic perspective to improve zero-shot action recognition accuracy with a multi-granularity semantic interaction mechanism. Inspired by ensemble learning, we propose a multi-level alignment (MLA) approach to compensate information for action classes. MLA aligns multi-granularity embeddings with visual embedding through a multi-head scoring mechanism to distinguish semantically similar action names and visually similar actions.
Furthermore, we introduce a new loss function sampling method to obtain a tight and robust representation. Finally, these multi-granularity semantic embeddings are synthesized to form a proper decision surface for classification. Significant action recognition performance is achieved when evaluated on the challenging NTU RGB+D, NTU RGB+D 120, and PKU-MMD benchmarks and validate that multi-granularity semantic features facilitate the differentiation of action clusters with similar visual features.  
\end{abstract}


\section{Introduction}
\begin{figure}[ht]
	\centering{\includegraphics[width=0.9\columnwidth]{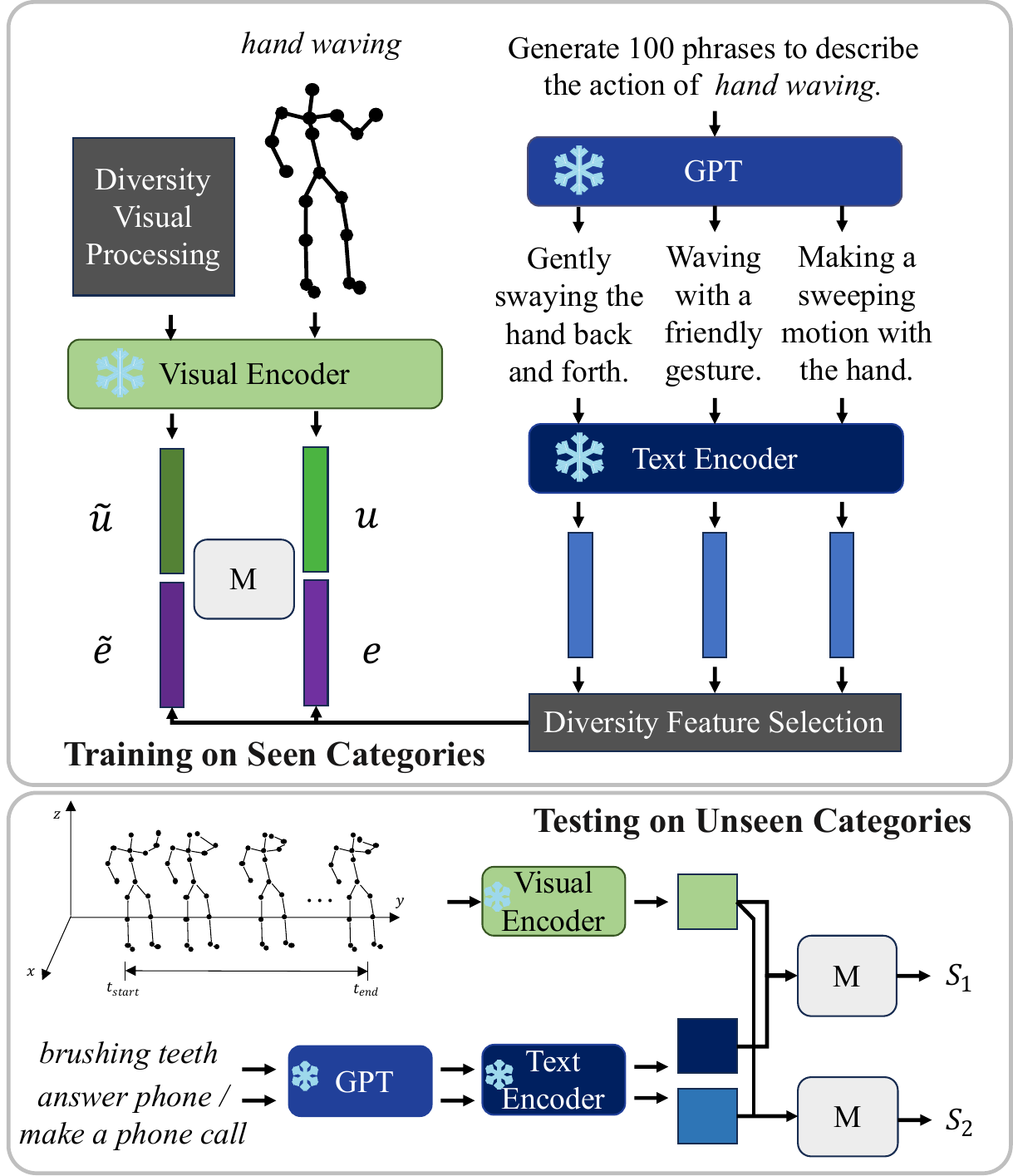}}
	\caption{The concept map for the zero-shot learning and our proposed method. During training, after the pre-trained language encoder extracts the description features, we let GPT generate diverse descriptions for each visible class, aligned with the visual features for multi-granularity alignment. When testing, the trained alignment network computes the similarity between the input samples and the unseen class description features to obtain the classification results.}
	\label{fig:concept_map}
\end{figure}
\IEEEPARstart{H}{uman} skeletal data has emerged as a promising alternative to traditional RGB video data due to its privacy-preserving nature. It is robust to appearance and background and can reflect an unbiased representation of group behavior. However, collecting and labeling many samples is not accessible due to the complexity of the data dimensions. Zero-shot learning (ZSL) \cite{ICCV_0012M019_RethinkZSL,cvpr_SchonfeldESDA19_CADAVAE}, in turn, is more flexible and economical in recognizing unknown categories. When the user wants to specify the category freely, the task can be modeled by modeling the connection between visual and semantic space in the seen category combined with an external knowledge base to obtain the semantic information of the specified category to reason about the desired result online.

The key to the problem of zero-shot skeleton-based human action recognition (ZSSAR) is to generalize the action concepts of categories from only known sequences and to recognize and explore a new unseen class properly.
Traditional approaches \cite{Gupta_2021_SynSE,Bhavan_2019_RelationNet} learn a compatible projection function \cite{Bhavan_2019_RelationNet} or depth metric \cite{nips_FromeCSBDRM13_DeViSE} for both visual and semantic space during the training phase. However, mapping features from different domains to a common anchor in the embedding space takes slight advantage of the semantic information. This means there is an assumption of low variability in the distribution of unknown and known classes in the feature space, leading to difficulties in generalizing to new classes with different distributions. Existing works \cite{Zhou_2023_SIME, Gupta_2021_SynSE} explore using textual embeddings to represent semantic information by passing the name of an action or a simple description to a pre-trained text encoder and computing it using manual prompts. Unfortunately, such single and simple semantic information is insufficient to achieve good zero-shot performance for skeleton modality, significantly reducing the general applicability of such zero-shot classifiers. 

\begin{figure}[t]
	\centering
	{\includegraphics[width=0.9\linewidth]{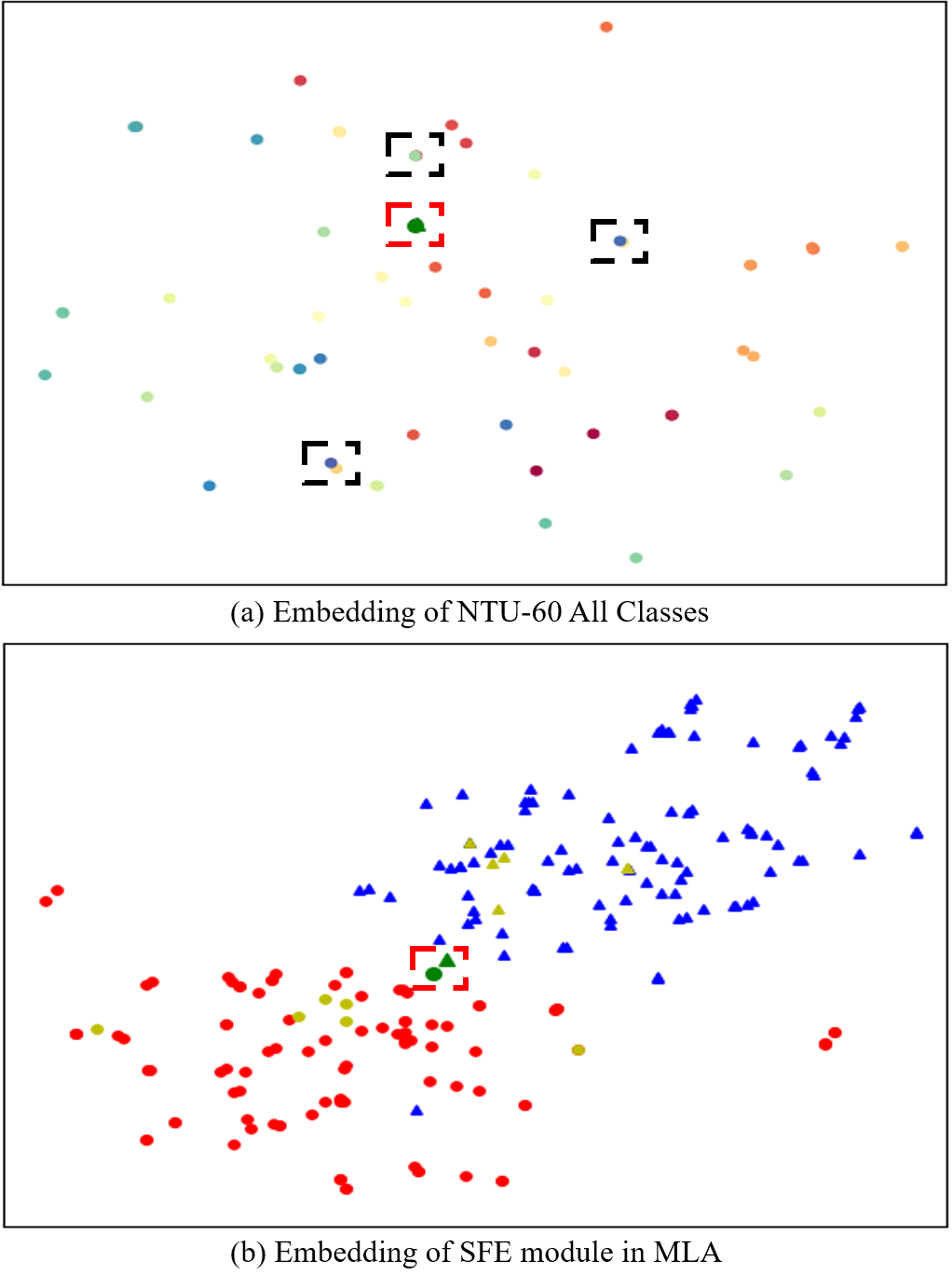}}
	\caption{We visualized the semantic features extracted from all categories in NTU 60 using t-SNE. The visualization in subfigure (a) shows cases where the class texts are similar, and the distance between their semantic features is relatively close. This can pose a potential confusion problem during testing. We visualize the aggregated semantic features of SFE modules in multiple branches for classes in subfigure (a) red dashed box, as shown in subfigure (b). Among them, circles and triangles indicate text features of classes with similar class names, respectively, the features in red and blue are the multiple description features of the two classes. The features in gold are the top-$ k $ features selected by SFE for them. The results show that in these two similar classes, compared to the original class description features, the margin of the multiple features aggregated by our SFE module is larger, corresponding to stronger separability.}
	\label{fig:m2visualization}
\end{figure}

In this paper, we design an information-compensation learning (InfoCPL) framework, as shown in Fig. \ref{fig:concept_map}, for the ZSSAR from an information-theoretic perspective. The framework enhances the model's ability to generalize and recognize new classes by enriching auxiliary information. It is noted that there is a double ambiguity of visual and semantic nature in skeleton data. First, some of the actions reflect too little difference in skeletal behavior, such as ``touch chest,'' ``stomachache,'' and ``heart pain,'' the visual embeddings are too similar to those of the other categories, and the visual and semantic embeddings are matched one-to-one. The simple descriptions' diversity is insufficient to draw the model's attention. Second, the discriminative power of the generated text embeddings is entirely dependent on the internal representation of the pre-trained text encoder. Some actions with similar class names are hard to distinguish in embedding space, as shown in Fig. \ref{fig:m2visualization}(a) red dashed box on NTU-60, ``walking towards each other'' and ``walking apart from each other.''

Specifically, this paper uses text descriptions to effectively complement the action names and the joint motion relationships for actions to capture the significant complexity of a given category.
First, we establish a codebook to automatically obtain sufficient visual descriptions by asking the large language model (LLM) questions such as ``Generate more phrases to describe the action of walking'' to obtain additional visual cues.
Based on the semantic codebook, we explore the expressiveness of text descriptions for whole sequences. Further, to enhance the separability of skeleton actions with minor visual differences and minor name differences, this paper proposes a multi-granularity semantic interaction mechanism, which is executed by a Multi-Level Alignment (MLA) module.
This component operates in parallel on skeleton sequences extracted from videos. Among other things, based on a simple cross-attention mechanism, the MLA is equipped with a Selective Feature Ensemble (SFE) module to obtain multi-granularity semantic embeddings from the codebook. To prevent the embedding collapse problem, we propose the attention inversion ($ A_{inv} $) mechanism to obtain a good initial distribution of semantic features.
Fig. \ref{fig:m2visualization}(b) visualization shows that MLA selects multi-granularity text description embeddings with reasonable inter-class distances in the semantic feature space.
Finally, inspired by ensemble learning, we evaluate the compatibility between a given visual representation of the skeleton and multi-granularity semantic text descriptions and joint motion descriptions using the ZSL component, i.e., learning a multi-head scoring network.
During training, paired visual features and multiple diverse semantic features constitute positive samples, and unpaired visual features and multiple semantic features constitute negative samples. 

In this paper, we explore the problem of how to align the set of linguistic descriptions with the set of skeleton samples in a better way for ZSL, and through the evaluation of the challenging NTU RGB + D 60 \cite{Shahroudy_2016_NTURGBD}, NTU RGB + D 120 \cite{Liu_2020_NTURGBD120}, and PKU-MMD \cite{LiuSLLH2020_PKUMMD} benchmarks, we show that 
(i) A generalized skeleton-based ZSL framework is proposed to identify and solve the ill-conditioned in generating semantic features using multi-granularity features for multi-level alignment and successfully improves the robustness of the model over multiple class splits;
(ii) A proposed codebook mechanism that utilizes visual features to access various semantic features generated by rich natural language descriptions emphasizes action-motion relationships, which outperforms previous work based on class name descriptions;
(iii) An attention inversion mechanism is proposed to create a distribution space with good diversity, improving feature learning quality;
Moreover, (iv) In order to achieve robust recognition of unknown actions, we modify the sampling of the loss function to incentivize the model to see more samples during training and achieve better results than the existing methods.

The remaining sections of this paper are organized as follows. First, we review related studies in Sec. \ref{sec:related_work}. Sec. \ref{sec:method} presents the methodology for solving the ZSSAR problem and the techniques for identifying significant motion regions in the skeleton. In Sec. \ref{sec:experiments}, we thoroughly evaluate and analyze the model. The full text is summarized in Sec. \ref{sec:conclusion}. We hope this work can help humans robustly deploy visual systems and effectively recognize unknown actions in a dynamic world while focusing on privacy preservation.

\section{Related Work}
\label{sec:related_work}
\subsection{Zero-Shot Action Recognition} 
There is a long tradition of research on ZSL, with origins dating back to \cite{aaai_LarochelleEB08}. On the other hand, zero-shot action recognition lies in successful recognizing unseen actions using knowledge learned during training \cite{Ye_2022_domainZSL,Gao_2020_CI_GNN}. This requires modeling the link between the visual feature and semantic representation spaces. Early models focused on using word embeddings as semantic representations, including action names \cite{cvpr_BrattoliTZPC20,cvpr_MandalNDGAK019,aaai_ShaoQL20}, defined action attributes \cite{cvpr_LiuKS11}, and descriptions of actions \cite{iccv_ChenH21,ICCV_0012M019_RethinkZSL}. Recently, the field has shown increasing interest in large language models \cite{eccv_NiPCZMFXL22,eccv_JuHZZX22,aaai_WuSO23}, using CLIP \cite{wang2021actionclip} and their variants to generate rich word embeddings for better visual language alignment. 
In contrast to these works, we focus on a more specific data modality: skeleton-based zero-shot action recognition. Compared to RGB videos, visual features of skeleton action sequences can be more challenging to utilize with less information.

\begin{figure*}[ht]
	\centering
	\includegraphics[width=\linewidth]{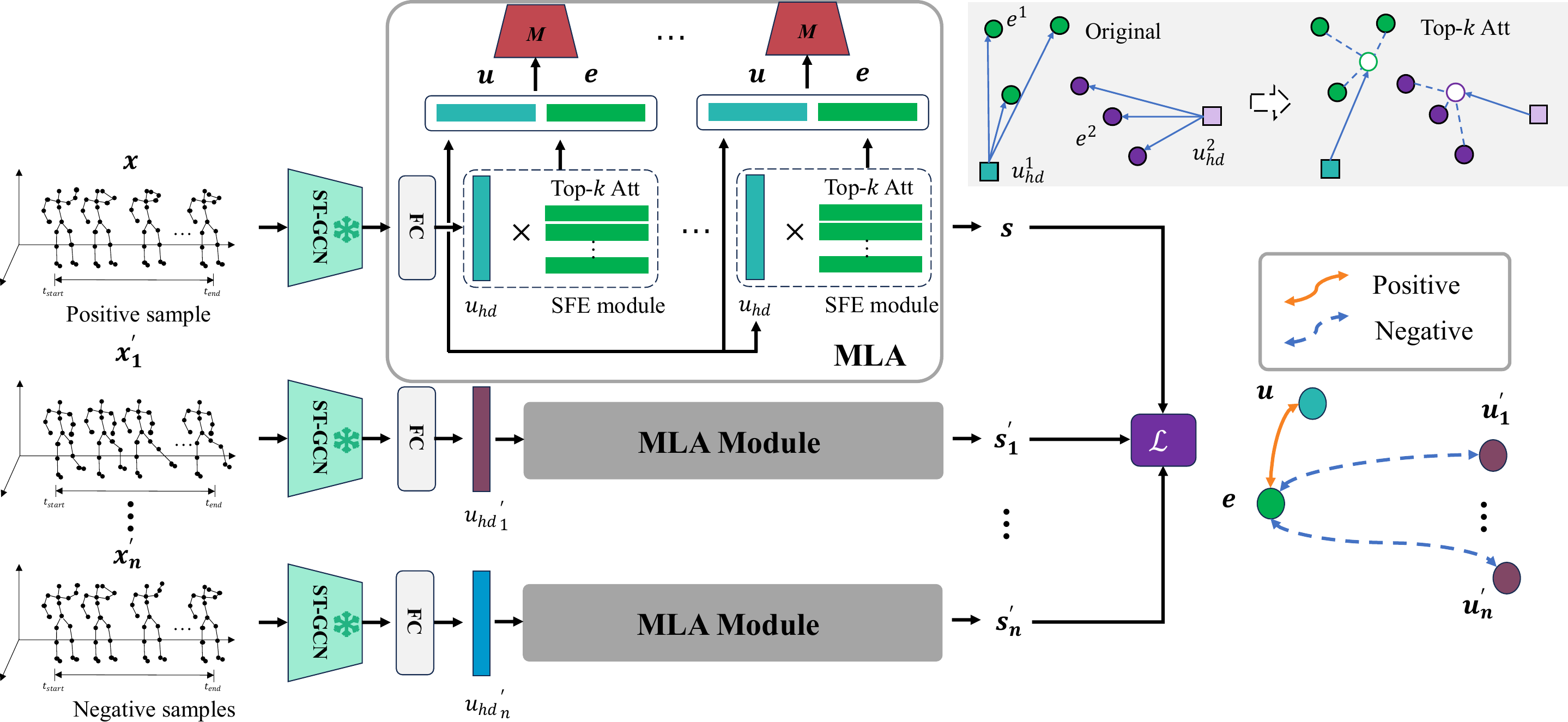}
	\caption{The pipeline of our framework. (a) Firstly, the positive example samples and negative example samples are pre-trained ST-GCN to extract the visual features. Then, multi-granularity scores are obtained after multi-head scoring averaging by our proposed MLA module, which consists of the Attention Inversion ($ A_{inv} $) strategy and the Multiple Semantic Feature Ensemble (SFE) module. In this, the $ A_{inv} $ strategy performs weight inversion to obtain an open semantic space for the semantic codebook composed of differentiated descriptive features at the early stage of training. The SFE selects from among the rich descriptive embeddings generated for each class to generate embeddings containing different granularities for alignment. Finally, MLA's multi-head scoring network synthesizes the multi-granularity embeddings in the semantic space to form decision surfaces with larger inter-class margin distance for classification. (b) Notation Summary: The visual features aligned with the high-dimensional semantic features are $ \boldsymbol{u_{hd}}$ and $\boldsymbol{{u_{hd}}'}$, which are positive and negative examples of the features, respectively. The similarity scores from the corresponding outputs of the MLA are scored as $ s $ and $ s' $ for the computation of the loss function. (c) A unique advantage with SFE modules. As shown in the upper right corner of the figure, SFE can synthesize multiple features of the same semantic meaning into anchor points based on visual-semantic cross-attention, using the enriched semantic codebook to get a better alignment effect.}
	\label{fig:framework}
\end{figure*}

\subsection{Skeleton-Based Action Recognition} 
In recent years, with the development of high-precision depth sensors and pose estimation algorithms such as the Kinect camera \cite{Cao_2017CVPR_2DPoseEstimation,ShottonSKFFBCM13_PoseEstiDepth}, human skeleton-based action recognition has received increasing attention due to its privacy-preserving nature. Human skeleton modalities \cite{aaa_LCLWD22_AugS,cvpr_LiWNWY021_CrossCLR} are robust to changes in the environment and object appearance, where each skeleton contains the major joints of the human body, each in its 2D or 3D coordinates. Conventional approaches project the skeleton to a pseudo-image \cite{CaetanoSBSS19_SkeleMotion} or connect as an input tensor \cite{Du_2015_Hierarchical} and utilize convolutional neural networks \cite{Li_2018_CNN} / recurrent neural networks \cite{pami_Zhang19_VALSTM}. The critical difference between ST-GCN \cite{Yan_2018_STGCN} and the previous ones is that it predefines a spatial topology map based on the natural connectivity of human joints and employs GCN to fuse skeletal joint information directly \cite{wang_2023_dearnet,wang_2023_ganet,xu_2023_maihar,Tian_2023_select}. On this basis, researchers employ more data streams or add attention mechanisms for modeling multi-level joint relationships. Notable examples include \cite{Shi_2019_2sAGCN,Cheng_2020_ShiftGCN,Chen_2021_CTRGCN,Chi_2022_InfoGCN,Ye_2020_DynamicGCN,Pang_2023_2sContrast}. More recently, the field has placed equal emphasis on the potential of large language models such as LST \cite{xiang2022language} and LA-GCN \cite{xu2023language}. In particular, LA-GCN combines the a priori knowledge of LLMs with the various processes of building graphs and reasoning and achieves state-of-the-art. However, the category description of LST and the a priori relation construction of LA-GCN are assigned to skeletons based on actual labels. In contrast, the ZSL framework does not require labeled samples for each action.

\subsection{Zero-Shot Skeleton-Based Action Recognition} 
The complexity of the semantics of human actions and the changing depth of information over time lead to difficulty labeling skeleton data, and researchers also have to consider the label noise influence in the study. Since ZSL allows models to recognize new actions based on their skeleton representation adaptively, this approach is crucial in the real world, as with zero-shot action recognition, existing skeleton-based approaches are based on building a mapping of visual and semantic features from the learning of seen classes in the training phase, and measuring the similarity between the visual features of the unseen classes and their sentence vectors or lexically labeled words to the learned common space in the testing phase. However, the above work needs to consider the existence of misalignment between visual and semantic feature distributions. 

Recently, large language models come to act as multimodal alignment modules to align visual features with parts of speech-tagged words. SynSE \cite{Gupta_2021_SynSE} requires part-of-speech (PoS) syntactic information to classify tags into verbs and nouns to enhance the generalization of the aligned model. SMIE \cite{Zhou_2023_SIME} aligns visual and textual embedding distributions by maximizing the mutual information between the two modalities and adds temporal constraints to enhance the model's performance further. Our approach differs from these works in two ways: We maximize the complementary information of the prompt to tailor an adequate description of the skeleton modality and, based on \cite{Zhou_2023_SIME}, ensure that the distribution between visual and semantic features can be aligned. Second, the use of richer textual semantics further emphasizes the fine-grained information of the skeleton sequence, allowing the model to learn from zero-shot while also having the ability to recognize fine-grained actions. It is worth noting that the above multimodal alignment modules have few fascinating presences of CLIP \cite{nips_Brown20_GPT3}. Because skeleton sequences, unlike natural images, cannot use the ready-to-use image-text alignment information in CLIP. Thus, prompt learning for LLMs is still a direction to further enhance the model's generalization ability. 


\section{Method}
\label{sec:method}
This section provides the foundational visual and semantic feature alignment framework for skeleton-based zero-shot action recognition (Sec. \ref{sec:pre}). Then, we elaborate on the proposed alignment approach using dense linguistic description construction (Multi-Level Alignment Module, Sec. \ref{sec:MLA}). Finally, an overview of the overall model's training approach and inference process is given (Sec. \ref{sec:train}).

\subsection{Preliminaries}
\label{sec:pre}
Models in the zero-shot setting need to be trained on visible classes and tested on disjoint unseen classes \cite{Bhavan_2019_RelationNet}. Each training sample in the training and testing datasets consists of a skeleton sequence and the corresponding class name. The skeleton sequence denotes $  \boldsymbol{x} \in \mathbb{R}^{ T \times V \times C} $, $ T $ is the length of frames, $ V $ is the number of joints, and $ C $ is the dimension of joint position coordinates. 

Typically, ZSL consists of a pre-trained visual feature extractor that has seen visible classes and a pre-trained large-scale language model as a semantic feature extractor. Then, visual and semantic features of skeleton sequences and class names are obtained by these two extractors. In other words, ZSL has two source domains: visual and auxiliary information domains. We use uppercase letters (e.g., $ U $ and $ E $) to denote random variables and lowercase letters (e.g., $ \boldsymbol{u} $ and $ \boldsymbol{e} $) to denote realizations of random variables $ U $ and $ E $, i.e., visual features $ \boldsymbol{u} $ and semantic features $ \boldsymbol{e} $, respectively.

Our goal is to learn a zero-shot classifier that can correctly assign test skeleton videos based on auxiliary semantic information. At test time, sequences are mapped twice for any category. A skeleton encoder $ g: X \to U $ maps unlabeled test sequences $ \boldsymbol{x} \in X $ to feature representations $ \boldsymbol{u} = g (X) \in \mathbb{R}^D $; and a scoring network $ h: (\mathbb{R}^{D'}, \mathbb{R}^{D'}) \to \mathbb{R}^{K} $ maps visual $ \boldsymbol{u} $ and semantic $ \boldsymbol{e} $ pairs to scores $ s \to R^K $, where $ D' $ and $ K $ denote feature dimensions and number of classes. Given a compatibility function $ M $, the test time zero-shot scoring function $ M: U \to S $ is defined as.
\begin{equation}
	\boldsymbol{s} = \text{argmax} \; M(\boldsymbol{u}, \boldsymbol{e}),
\end{equation}
where the structure of $ M $ is a MLP layer. 
In brief, the performance of the ZSSAR model is directly dependent on three factors:(i) the modal representation of the skeleton, (ii) the diversity semantic representation, and (iii) the learning model of the scoring function $ M $. We solve the ZSSAR problem based on the proposed multi-granularity semantic information interaction mechanism, with details presented in the following sections. 

\subsection{Framework}
\label{sec:framework}
Our framework is shown in Fig. \ref{fig:framework}. First, the pretrained ST-GCN is the encoder network performing on positive samples and negative samples to extract visual features. Then, the proposed Multi-Level Alignment (MLA) module acquires multiple multi-granularity semantic embeddings and obtains multiple scores with the visual features after multi-head scoring. Ultimately, averaging these scores yields a multi-granularity score for that skeleton sequence. 
Specifically, we concatenate the visual feature $ \boldsymbol{u} $ into a positive pair with the paired semantic feature $ \boldsymbol{e} $ and concatenate that $ \boldsymbol{e} $ into a negative pair with other sequences of visual features $ \boldsymbol{u'} $. These two kinds of pairs are input into estimation networks $ M $ to obtain scores $ s $ and $ s' $ for contrastive learning to get the multi-level alignment.  In the following, we describe the implementation details and ideas of the MLA module.

\subsection{Multi-level Alignment Module} 
\label{sec:MLA}
To complement ZSL, we propose a multi-level alignment (MLA) module based on rich descriptions to ensure diversity in the semantic features provided at each level. The MLA is inspired by the ensemble learning \cite{random_subspace,Random_Forests}, whose function is to realize a method that combines the prediction results of multiple models (the scoring network $ M $) to obtain a more robust prediction than any of the models. 
As shown in Fig. \ref{fig:framework}, the MLA module consists of $ n $ branches, and each level generates one for matching $ \boldsymbol{e} $ for the visual feature $ \boldsymbol{u} $. The MLA average the prediction scores $ s $ of the estimation module $ M $ of each branch for $ (\boldsymbol{u}, \boldsymbol{e}) $ as the final prediction scores $ s_{multi} $ of the current class. 

In this case, we propose the semantic feature ensemble (SFE) module as shown in Fig. \ref{fig:framework}. The function of SFE is to ensure each scoring network in MLA uses embeddings with different granularity, reducing the correlation between each scoring network to prevent the failure of ensemble learning \cite{random_subspace}. In addition, the SFE module augments the action labels to satisfy the need for semantic information. We introduce the SFE module in detail in the following subsection.

\subsubsection{Single Branch in MLA}
\label{sec:single_branch}
The implementation of the SFE module is based on text augmentation to select more disparate semantic features of different granularity for visual feature alignment. It is positioned between the visual coder and the estimation module $ M $. For each class of visual features, we first use the LLM to generate paired descriptions for them to extract semantic features. The visual feature $ \boldsymbol{u} $ and text feature $ \boldsymbol{e'} $ of the corresponding category are sent as inputs to the SFE module. 

Taking a branch as an example, the visual feature $ \boldsymbol{u} $ first goes through a projection layer to get $ \boldsymbol{u_{hd}} $ and maps the feature dimensions to the same number as $ \boldsymbol{e'} $. In each branching level, we select the most similar top-$ k $ semantic features among all the semantic features of the class through the SFE module to generate attention for generating new semantic features. The value of $ k $ increases accordingly with the increase of branching. Finally, $ n $ semantic features with different granularity are generated for alignment where top-$ k $ attention is the following form.
\begin{equation}
	\Phi(\boldsymbol{e'},\boldsymbol{u_{hd}}) = sort_k(\frac{\boldsymbol{e'} \times \boldsymbol{u_{hd}}}{\sqrt{C_e}}),
\end{equation}
where key = value = $ \boldsymbol{e'} $ (semantic feature), the query is a visual feature, $ C_e $ is the feature dimension of $ \boldsymbol{e'} $, $ sort_k $ is the top-$ k $ selection operation, it sets the attention scores not in the top-$ k $ to $-\infty$. Then, the $ \boldsymbol{e'} $ is weighted after softmax and used as the output of SFE: $ \boldsymbol{e} = \text{softmax}(\Phi(\boldsymbol{e'},\boldsymbol{u_{hd}})) \times \boldsymbol{e'} $. 
Ultimately, this output feature is the $ \boldsymbol{e} $ in that branch aligned to $ \boldsymbol{u} $.  
During inference, the estimation network $ M $ in each branch scores all $ (\boldsymbol{u}, \boldsymbol{e'}) $ pairs and then averages them as the score for each unseen category.

\subsubsection{Semantic Embedding Codebook}
The 100 description embeddings constructed manually can be considered a finite number of semantic codebooks for skeleton visual embeddings. SFE quantizes visual features into a semantically discrete space by embedding visual feature vectors into the top-$ k $ embeddings with the maximum attention within the semantic codebook. This process can be defined as visual vector quantization (VQ). It is a fundamental component of many representation learning techniques in machine learning by reducing the search space to store domain information in compact representations vital in the latent space. Specifically, we first want to get as many textual descriptions as possible for a class name. Referring to the observation in \cite{ICML_allingham23a_ZPE,icml_KaulXZ23}, the final performance when using a large number of cues to extract semantic features has a long-tailed distribution as the number of cues increases, with the empirical number of decays ranging from 50-100. Therefore, we require an autoregressive language model to generate 100 descriptions for each class name. In other words, we input the GPT \cite{nips_Brown20_GPT3} question prompt as ``Generate 100 phrases to describe the action of \{class name\}.'' Then, text features are extracted for the 100 descriptions, and each sentence text feature can be represented as $ \boldsymbol{e'} \in \mathbb{R}^{100 \times C_e} $.

We have tried to provide richer hierarchical information for semantic features with the MLA. Meanwhile, inspired by the multi-stream nature of the data (topologies modality \& motion modality) in traditional 3D skeleton-based action recognition, we use the motion information about the actions to enhance the visual features further and help the model to enhance the execution regions of interest for different actions. However, to our surprise, GPT rejected our request to generate 100 motion descriptions. To this end, we extract features for the unique motion description. After normalization, $ \boldsymbol{m} \in \mathbb{R}^{1 \times C_e} $ is directly added with the output  $ \boldsymbol{e} $ of the SFE in Sec. \ref{sec:single_branch} to get the complete description feature. We are then aligned with the visual feature $ \boldsymbol{u} $.
 
\subsubsection{Critical Attention Inverse}
A question exists here whether multi-branch top-k is meaningless if only maximal attention is learned. The answer we give is yes. At the beginning of MLA design, the purpose is to find semantic features with considerable diversity for alignment. On the contrary, if the maximum attention is used from the beginning to the end, its performance will be weakened as the iteration progresses. This is because, under the attention mechanism, only a small fraction of the embeddings are active, while most embedding vectors are snowy and will never be used. In other words, the semantic embeddings used for optimization become homogenization. 

To address the codebook collapse problem, we propose a new inverse attention mechanism $ A_{inv} $ to reverse the dynamic initialization of the embedded attention in the semantic codebook at the initial stage of training. The method is inspired by previous dynamic cluster initialization techniques \cite{ReInitial_KMeans}, where the code vectors are resampled to obtain a new initialization. There are two reasons for $ A_{inv} $ proposed. First, the embeddings in the codebook remain the same semantic as the visual embedding; even the lowest-attention embeddings fill the bill. Moreover, we need more separable features, i.e., by choosing embeddings with a greater distance in the feature space to improve inter-class separability. Specifically, in the initial stage of training, we replace the Top-$ k $ Att in the SFE module of each branch in MLA with inverted attention. Meanwhile, all other settings are kept unchanged. The mechanism can be represented as follows:
\begin{equation}
	\Phi(\boldsymbol{e'},\boldsymbol{u_{hd}})_{start} = - \Phi(\boldsymbol{e'},\boldsymbol{u_{hd}}), \; if \; epoch < N_{ep}.
\end{equation}
where $ N_{ep} $ is a set value where we use the same number of steps as warming up. It allows the MLA model to gradually disrupt the distribution of the original semantic space using reverse top-$ k $ attention following iterations when the epoch is smaller than this value. At this stage, the visual features de-align the less relevant descriptions to obtain a good initial semantic space.

\subsubsection{Discussion}
Since different class divisions significantly impact the ZSL results, even if the number of unseen classes is set to be the same, the final performance of the model can be highly biased. Zero-shot skeleton action recognition provides a three splits test experimental design \cite{Zhou_2023_SIME} to increase the confidence of the results. In selecting semantic features, we observe that directly averaging all 100 semantic features as the final $ \boldsymbol{e} $ to match with $ \boldsymbol{u} $ already yields good performance. However, this approach excels in one division and performs mediocrely in the other two. To enhance the model robustness of ZSL, MLA branches can perform well on all splits, as shown in Fig. \ref{fig:tsne} and \ref{fig:visualization}.

\subsection{Training Overview.}
\label{sec:train}
We train this framework using contrastive self-supervised loss \cite{He_2020_CVPR,hjelm2018learning}. Let $ i $ be the index of the current positive sample and $ j $ be the index of the other negative samples. The loss takes the following form:
\begin{equation}
	\label{eq:nce}
	\begin{aligned}
		L = - {\frac{1}{K}}\sum_{i=1}^{K}\log{\frac{e^{f(x_{i},y_{i})}}{\sum_{j=1}^{K}e^{f(x_{i},y_{j})}}},
	\end{aligned}
\end{equation}
where $ f(x_{i},y_{i}) = f_{sp}(M(\boldsymbol{u},\boldsymbol{e})) $, $ f(x_{i},y_{j}) =  f_{sp}(M(\boldsymbol{u},\boldsymbol{e'})) $, $ M $ is scoring networks, and $ f_{sp} $ is the soft-plus function $ f_{sp}(z) = log(1 + e^z) $. $ (\boldsymbol{u},\boldsymbol{e}) $ is the visual and semantic features of the uniform class. $ (\boldsymbol{u},\boldsymbol{e'}) $ is a negative sample pair because $ \boldsymbol{e'} $ is a semantic feature related to other classes. 

The form of the sum of negative examples in the denominator of Eq. \ref{eq:nce} suggests that the learning of different classes of skeleton distributions is driven to a large extent by noise-contrast estimation \cite{gutmann2010nce,Sohn_2016_Npair}. In other words, representation learning is done by adding more negative examples to improve the model's ability to distinguish between signal (positive sample pairs) and noise (negative sample pairs). Much work has demonstrated that models perform better as negative data increases \cite{Tian_2020_what,Bachman_2019_Learning}. In addition, this paper emphasizes improving the model's generalization ability by compensating for information and increasing feature variability in the ZSSAR task. Thus, we argue that the sampling way limits the current loss version. Specifically, the input samples are sampled based on their labeling information $ y_j $. The number of labels is highly correlated with the dataset; for example, the sampling space of NTU-60 has only 59 possibilities, and the available samples are even fewer, considering the number of samples within a batch. However, if the different skeleton sequences $ x_j $ are sampled directly, the sampling space expands to the entire batch. For this purpose, we used the following training loss:
\begin{equation}
	\label{eq:nce2}
	\begin{aligned}
		L = - {\frac{1}{K}}\sum_{i=1}^{K}\log{\frac{e^{f(x_{i},y_{i})}}{\sum_{j=1}^{K}e^{f(x_{j},y_{i})}}},
	\end{aligned}
\end{equation}
where $ f(x_{j},y_{i}) =  f_{sp}(M(\boldsymbol{u'},\boldsymbol{e})) $, $ (\boldsymbol{u'},\boldsymbol{e}) $ is a negative sample pair because $ \boldsymbol{u'} $ is a visual feature related to other classes. In summary, the loss has the following advantages: (1) more flexible for X-sampling, not limited by the number of dataset categories. (2) better diversity of negative samples for X-sampling.

In the inference phase, the trained estimation network $ M $ computes the similarity scores $ s $ between the visual sequences in the test set and the semantic features of all unseen classes. The unseen class with the highest similarity score is selected as the prediction for that test sequence.

\section{Experiments}
\label{sec:experiments}

In this section, we first describe the dataset used, implementation details, and baseline settings, and then give and discuss the results of our experiments on ZSSAR. 

\begin{table}[t]
	\caption{Comparison of InfoCPL with the State-of-the-Art methods on NTU-60 and NTU-120 datasets. Splits provided by SynSE.}
	\label{tab:ntu_60_120}
	\centering
	\begin{tabular}{|l||c||c||c||r|}
		\hline
		Method & \multicolumn{2}{c||}{NTU-60} & \multicolumn{2}{c|}{NTU-120}  \\
		\hline
		split & 55/5 & 48/12 & 110/10 & 96/24 \\
		\hline
		DeViSE      & 60.72 & 24.51 & 47.49 & 25.74 \\
		RelationNet & 40.12 & 30.06 & 52.59 & 29.06 \\
		ReViSE      & 53.91 & 17.49 & 55.04 & 32.38 \\
		JPoSE       & 64.82 & 28.75 & 51.93 & 32.44 \\
		CADA-VAE    & 76.84 & 28.96 & 59.53 & 35.77 \\
		SynSE       & 75.81 & 33.30 & 62.69 & 38.70 \\
		SMIE        & 77.98 & 40.18 & 65.74 & 45.30 \\
		\rowcolor{lightgray} MSF-GZSSAR      & 83.63 & 49.19 & 71.20 & 59.73 \\
		\hline
		\textbf{InfoCPL}   & \textbf{85.91} & \textbf{53.32} & \textbf{74.81} & \textbf{60.05} \\
		\hline
	\end{tabular}
\end{table}

\subsection{Benchmarks and Implementation Details}
\subsubsection{Datasets} 
In this work, most of the experiments are performed based on three skeleton-based action recognition datasets NTU-RGB+D 60 \cite{Shahroudy_2016_NTURGBD} (NTU-60), NTU-RGB+D 120 \cite{Liu_2020_NTURGBD120} (NTU-120), and PKU-MMD \cite{LiuSLLH2020_PKUMMD}, which contain a large number of daily, health, and interaction behaviors. Specifically, NTU-60 contains 56,578 skeleton sequences for 60 movement categories. Human skeletal sequences were captured using a Microsoft Kinect v2 sensor, represented by 25 joint points per subject. NTU-120 is an extended version of NTU 60, containing 114480 skeletal sequences across 120 action categories, the largest dataset used for action recognition. PKU-MMD captured 51 categories, including nearly 20000 samples. The dataset consists of two parts (21,539 / 6,904 samples), and the second part is more challenging with a greater variation of viewpoints.

\begin{table}[t]
	\caption{The average accuracy of 3 class splits on NTU-60, NTU-120 and PKU-MMD.}
	\label{tab:ntu_60_120_pku}
	\centering
	\begin{tabular}{|l||c||c||c|}
		\hline
		Method & NTU-60 & NTU-120 & PKU-MMD  \\
		\hline
		split & 55/5 & 110/10 & 46/5  \\
		\hline
		DeViSE      & 49.80 & 44.59 & 47.94 \\
		RelationNet & 48.16 & 40.55 & 51.97 \\
		ReViSE      & 56.97 & 49.32 & 65.65 \\
		SMIE        & 63.57 & 56.37 & 67.15 \\
		SMIE\_Chat  & 70.21 & 58.85 & 69.26 \\
		\hline
        \textbf{InfoCPL}   & \textbf{80.96} & \textbf{70.07} & \textbf{85.15} \\
		\hline
	\end{tabular}
\end{table}

\subsubsection{Implementation Details} 
\begin{figure}[ht]
	\centering
	{\includegraphics[width=0.8\linewidth]{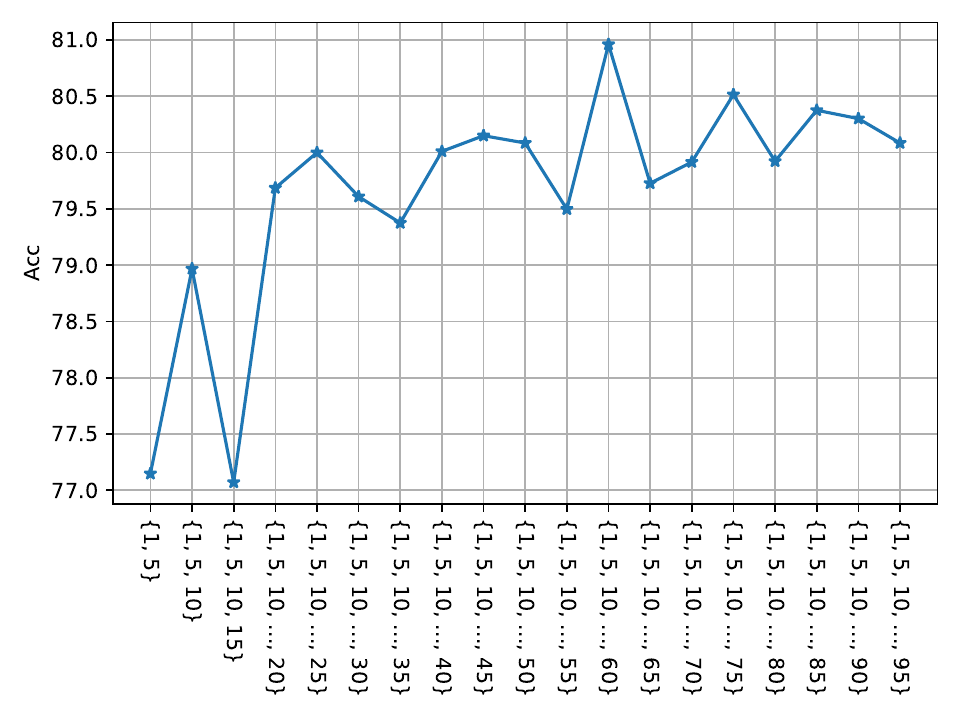}}
	\caption{Comparisons of different $ k $ for top-$ k $ attention in MLA on NTU-60 datasets. The interval for $k$ is 5; the maximum number of branches is 20.}
	\label{fig:top_k}
\end{figure}

The training phase followed the same data processing procedure used in Cross-CLR \cite{cvpr_LiWNWY021_CrossCLR} and SMIE \cite{Zhou_2023_SIME}, with all skeleton sequences resized to 50 frames by linear interpolation. For text features, GPT \cite{nips_Brown20_GPT3} was used to generate extensive paragraph descriptions for the actions, and Sentence-Bert \cite{EMNLP_ReimersG20_SentenceBert} was used to obtain 768-dimensional word embeddings for each sentence in the paragraphs. Then, $ \text{L}_2 $ normalization was applied to all features for training stability. Meanwhile, ST-GCN \cite{Yan_2018_STGCN} and Shift-GCN \cite{Cheng_2020_ShiftGCN} pre-trained models are used as the backbone to extract visual features with a hidden channel size of 16, and the final extracted features have a dimension of 256. Then, a fully connected layer is used to project the features from 256 to 768 dimensions. The estimation network $ M $ consists of three MLP layers with ReLU \cite{Nair_2010_ReLU} (MLP channels: $ 1536 \to 1024 \to 512 \to 1 $). We trained the network on an NVIDIA RTX 3090 GPU with an Adam optimizer and a 100 epoch CosineAnnealing scheduler with a batch size 128. The learning rate was $ 1e - 5 $  on NTU-60 and PKU-MMD; on NTU-120 is  $ 1e - 4 $. The learning rate warming up and $ A_{inv} $ warming up are both 15 epochs.
Meanwhile, $ k $ in top-$ k $ attention of the SFE module is 1\_60\_5 (\{1, 5, 10, $ \cdots $, 60\}, see \ref{fig:top_k}) on NTU-60, 1\_25\_5 on NTU-120, 1\_15\_5 on PKU-MMD. 

\begin{table}[t]
	\caption{The ablation study of InfoCPL. Loss: Use Xsample InfoNCE; MD: Use the multi-description codebook; Agg: Aggregation method of MD embeddings in one branch; Avg: Average score fusion; Att: Attention-based score fusion; Top-k: Top-k attention score fusion; MLA: Multi-level alignment; $A_{inv}$: Critical attention inverse.}
	\label{tab:ex_description}
	\centering
	\begin{tabular}{|l||c||c||c||c||c||c|}
		\hline
		& \multicolumn{5}{c||}{Components} & \multirow{2}{*}{NTU-60(\%)}                                             \\
		\cline{2-6}
		& Loss                      & MD              & Agg           & $A_{inv}$    & MLA           &       \\
		\hline
1) & $\checkmark$                   &                 &               &              &               & 71.72 \\
\hline
2) & $\checkmark$                   & $\checkmark$    & Avg           & --           &               & 74.41 \\
3) & $\checkmark$                   & $\checkmark$    & Att           &              &               & 76.93 \\
4) & $\checkmark$                   & $\checkmark$    & Top-k         &              &               & 76.12 \\
\hline
5) & $\checkmark$                   & $\checkmark$    & Avg           & --           &               & 74.41 \\
6) & $\checkmark$                   & $\checkmark$    & Att           & $\checkmark$ &               & 77.01 \\
7) & $\checkmark$                   & $\checkmark$    & Top-k         & $\checkmark$ &               & 77.83 \\
\hline
8) & $\checkmark$                   & $\checkmark$    & Top-k         & $\checkmark$ & $\checkmark$  & \textbf{80.96} \\
		\hline
	\end{tabular}
\end{table}


\subsubsection{Baseline Methods} 
We use seven zero-shot learning methods based on modeling visual and semantic space connections as a baseline. They are DeViSE \cite{nips_FromeCSBDRM13_DeViSE}, ReViSE \cite{iccv_TsaiHS17_ReViSE}, RelationNet \cite{Bhavan_2019_RelationNet}, JPoSE \cite{Iccv_WrayCLD19_JPoSE}, CADAVAE \cite{cvpr_SchonfeldESDA19_CADAVAE}, SynSE \cite{Gupta_2021_SynSE}, and SMIE\cite{Zhou_2023_SIME}, which maps visual and semantic features into the same space. DeViSE uses a dot product, which computes the similarity between the mapped features. RelationNet designs a relational module. ReViSE uses maximum mean difference loss to align two-part embeddings. JPoSE learns lexically aware embeddings and constructs a separate multimodal space for each lexical label to perform cross-modal fine-grained action retrieval between text and skeleton. CADA-VAE learns visual and semantic features by variational autoencoders (VAE) in the alignment space. On this basis, SynSE trained multiple VAEs to incorporate lexical and syntactic information into potential visual representations of the skeleton. On the other hand, SMIE globally aligns the visual and semantic spaces using mutual information maximization and employs temporal constraints for regularization. We provide a comprehensive comparison between InfoCPL and these baseline approaches.


\subsection{Evaluation and Comparison}

\subsubsection{Compare with SOTA Methods}
Table \ref{tab:ntu_60_120} shows the results of our work on NTU-60 and NTU-120, which uses textual cue-based integration to extract semantic features based on visual sequence augmentation to increase feature diversity, compared to a series of previous baseline. For fair evaluation, we report two fixed class divisions on the NTU-60 and NTU-120 datasets \cite{Gupta_2021_SynSE}. The gray result in Table \ref{tab:ntu_60_120} indicates that text features were extracted using CLIP-text encoder, and the rest are extracted text features using Sentence-BERT. For NTU-60, samples were divided into two settings of 55/5 and 48/12, and for NTU-120, 110/10 and 96/24. In other words, using the 110/10 division as an example, we validated the model using 110 visible classes and 10 unseen classes, with the visual feature extractor Shift-GCN \cite{Cheng_2020_ShiftGCN} and the semantic feature extractor machine Sentence-Bert \cite{EMNLP_ReimersG20_SentenceBert}.
For both splits on the NTU-60 dataset, InfoCPL outperforms SMIE by 7.93\% and 12.11\%, respectively. For NTU-120, the InfoCPL reached 9.07\% and 7.75\%. GZSSAR \cite{li2023multisemantic} is the method using a CLIP-text encoder as shown in the gray part of Table \ref{tab:ntu_60_120}. CLIP was trained using 400 million pairs of data searched online, whereas Sentence-BERT was trained using only about 1.8 million pairs. Specifically, GZSSAR directly aggregates the three specific descriptions into semantic embeddings (action names, action descriptions, and motion descriptions), maps the visual and semantic embeddings to their separate spaces, and aggregates the visible and invisible class classifiers to obtain the final results. With a weaker feature extractor than CLIP, InfoCPL outperforms GZASSAR by 2.28\% and 4.13\% on both splits of NTU-60, with 3.61\% improvement on NTU-120 110/10 split and comparable on the 96/24 split. The results show that ensuring that the model sees more diverse visual distributions during ZSL classifier training facilitates the formation of a good representation space without the need to map the embeddings separately. In addition, another critical point is that using more discriminative semantic embeddings consistent with the visual embeddings can help the model form suitable decision surfaces. As the number of unseen classes increases, the difficulty of predicting unseen classes increases, and the proposed InfoCPL method utilizes semantic cues ensemble to improve the model's potential in the face of unknown classes.

\subsubsection{Results on Tri-Splits Protocol}
Table \ref{tab:ntu_60_120_pku} shows the further optimized experimental setting for zero-shot skeleton action recognition, where the dataset was extended from two large-scale skeleton datasets to three large-scale skeleton datasets, i.e., NTU-60, NTU-120, and PKU-MMD, to mitigate the effect of different class divisions on the results. We refer to SMIE for each dataset to perform a three-splits test to reduce the result variance. Different seen and unseen categories are set for each fold, and the average results are reported as the standard. Under this test, we compare our work with models that use the same experimental setting.
DeViSE and RelationNet are based on projection, and ReViSE and SMIE are based on distributional alignment.
In all cases, these models apply the classical ST-GCN \cite{Yan_2018_STGCN} as a visual feature extractor to control its impact on the experiment. As with SMIE \cite{Zhou_2023_SIME}, we used Sentence-Bert \cite{EMNLP_ReimersG20_SentenceBert} as a semantic feature extractor for evaluation.
The visible/unseen splits on the three datasets are 55/5, 110/10, and 46/5, respectively.
It can be observed that the projection-based approach obtains lower accuracy than the domain-aligned approach and that InfoCPL achieves significant improvements on all datasets. InfoCPL outperforms other projection-based methods on the three datasets and achieves 10.75\%, 11.22\%, and 15.89\% compared to other alignment methods.
These results indicate that the global and detailed alignments generated by our method using rich class descriptions provide more favorable additional information.

\subsection{Ablation Study}
\subsubsection{Loss Function}
We evaluate three widely used ZSL loss function methods, i.e., SoftmaxCE \cite{liu2016large}, JSD \cite{nips_NowozinCT16_JSD}, and InfoNCE \cite{He_2020_CVPR}. In this comparison, we score single visual and semantic embeddings on the NTU-60 dataset using only one scoring network $ M $ in Sec. \ref{sec:pre} for simplicity. We use a tri-fold validation protocol to compare loss performance.

The results are shown in Table \ref{tab:loss_func_exp}, with T being the temperature parameter experiment for SoftmaxCE. infoNCE loss has the best performance, demonstrating the effectiveness of using more samples for training. Both JSD and InfoNCE maximize the mutual information between the two stochastic variables, and their poor performance may be because their fixed ratio of positive and negative samples (2:1) restricts the model's performance. Since the effective data augmentation of the skeleton data is small, it cannot provide enough positive samples, resulting in sub-optimal model performance. On the contrary, our results using InfoNCE to directly sample $ X $ are displayed in Table \ref{tab:infonce_sample}. The results for `Xsample' are highest at a batch sampling number of 32, with 71.78\% accuracy. However, considering the energy efficiency ratio, we chose a sampling number of 8 for subsequent experiments.

\begin{table}[t]
	\caption{Experiments with different loss function on NTU-60. T is the temperature parameter for SoftmaxCE. When T less than 1, the function produces a sharp distribution and vice versa. The default temperature parameter used by MoCo is 0.07 (logit divided by T, then softmax).}
	\label{tab:loss_func_exp}
	\centering
	\resizebox{\linewidth}{!}{
		\begin{tabular}{|c||c||c||c||c||c|}
			\hline
			Loss & \multicolumn{5}{c|}{Avg Acc} \\
			\hline
			\multirow{2}{*}{SoftmaxCE} & T = 2 & T = 1 & T = 0.5 & T = 0.1 & T = 0.07  \\
			\cline{2-6}
			       & 68.30 & 68.44 & 68.60 & 67.95 & 67.72 \\
			\hline
			JSD    &  \multicolumn{5}{c|}{69.28} \\
			\hline
			InfoNCE  & \multicolumn{5}{c|}{\bf 70.59} \\
			\hline
	\end{tabular}}
\end{table}


\begin{table}[t]
	\caption{Experiments for InfoNCE builds negative samples on NTU-60. The activation function defaults to softplus.}
	\label{tab:infonce_sample}
	\centering
	\resizebox{\linewidth}{!}{
		\begin{tabular}{|l||c||c||c||c||c||c|}
			\hline
			Numbers & 1 & 2 & 4 & 8 & 16 & 32 \\ 
			\hline
			Ysample & \bf 70.59 & 69.91 & 69.74 & 69.01 & 68.85 & 68.57 \\
			Xsample & 70.00 & 70.21 & 70.87 & \underline{71.72} & 71.53 & \textbf{71.78} \\
			\hline
	\end{tabular}}
\end{table}

\subsubsection{Expanding Category Descriptions for More Information}
The ablation experiments for the proposed InfoCPL are reported in Table \ref{tab:ex_description}. 
In Table \ref{tab:ntu_60_120_pku}, SMIE takes the action name as input and has 63.57\%. SMIE\_Chat for making GPT expand the action name into a full description and has 70.21 \%. In Table \ref{tab:ex_description}, "MD" is the average feature performance using only the generated description of 74.41 \%. We input the GPT once (a sentence containing the action name); however, the information obtained was more, indicating that the prompt design was successful and did not require time-consuming selection through prompt engineering.

Further, rows 2) to 4) of Table \ref{tab:ex_description} evaluate the effect of multiple descriptions using a different aggregation method, “Agg,” and compare it with the MD direct averaging scheme. Where ``Att'' denotes the choice of semantic embedding using global attention and ``Top-$ k $'' denotes the use of top-$ k $ attention. For all results, we still use only one $ M $ for scoring.
As can be seen, the performance of using top-$ k $ attention directly is weaker than that of using global attention. As mentioned earlier, the semantic codebook collapses. With iterative optimization models tend to update the weights of only the first few semantic embeddings thus affecting model generalization.

\subsubsection{Critical Attention Inverse in the Initialization Stage}
\begin{figure}[b]
	\centering
	{\includegraphics[width=0.7\linewidth]{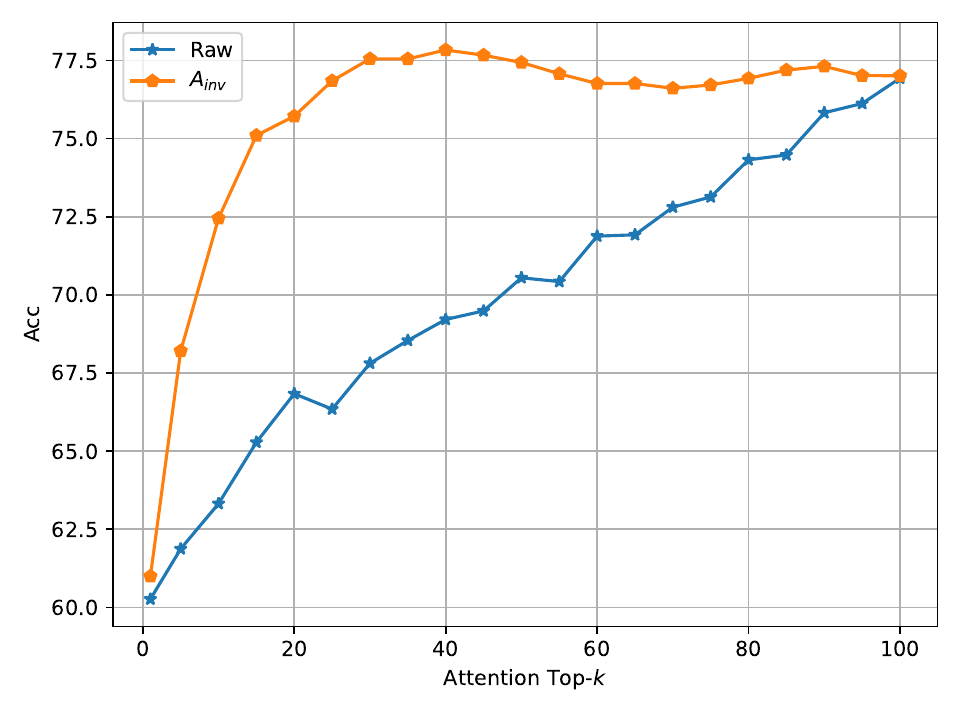}}
	\caption{Intuitive visualization of $ A_{inv} $ improvement on the single branch.}
	\label{fig:ainv_effect}
\end{figure}

Rows 5) to 7) of Table \ref{tab:ex_description} evaluate the capabilities of $ A_{inv} $.
When trained with $ A_{inv} $, the semantic embeddings selected by SFE are farther away from the visual embeddings, with a performance improvement of 3.42 \% in Table \ref{tab:ex_description}, which facilitates the recognition of similar action name categories. Furthermore, comparing the performance of MDs with different aggregations before and after the addition of $ A_{inv} $ shows that the top-$ k $ mechanism becomes successful because $ A_{inv} $ disrupts the distribution of semantic embeddings.
The orange curve in Fig. \ref{fig:ainv_effect} demonstrates the significant boost from adding $ A_{inv} $. The horizontal axis is the change in the value domain of $ k $, corresponding to the single branch performance using different top-$ k $ attention.
This result suggests that the top-k mechanism increases the diversity of semantic embeddings.

\begin{table}[t]
	\caption{Experiments on the generalizability of MLA over the ZSSAR algorithm on NTU-60.}
	\label{tab:general_mla}
	\centering
	\resizebox{\linewidth}{!}{
		\begin{tabular}{|c||c||c||c||c||c||c||c|}
			\hline
		    \multicolumn{6}{|c||}{NTU-60 ZSL} & \multicolumn{2}{c|}{ImageNet ZSL} \\
			\hline
			\multicolumn{2}{|c||}{DeViSE} & \multicolumn{2}{c||}{RelationNet} & \multicolumn{2}{c||}{ReViSE} &\multicolumn{2}{c|}{CuPL} \\
			\hline
			 w/o & w & w/o & w & w/o & w & w/o & w \\ 
			\hline
			49.8 & \textbf{51.87} & 48.16 & \textbf{52.04} & 56.97 & \textbf{59.91} & 76.69 & \textbf{76.88} \\
			\hline
	\end{tabular}}
\end{table}

\subsubsection{Multi-level Alignment Module}
\begin{figure}[b]
	\centering
	{\includegraphics[width=0.7\linewidth]{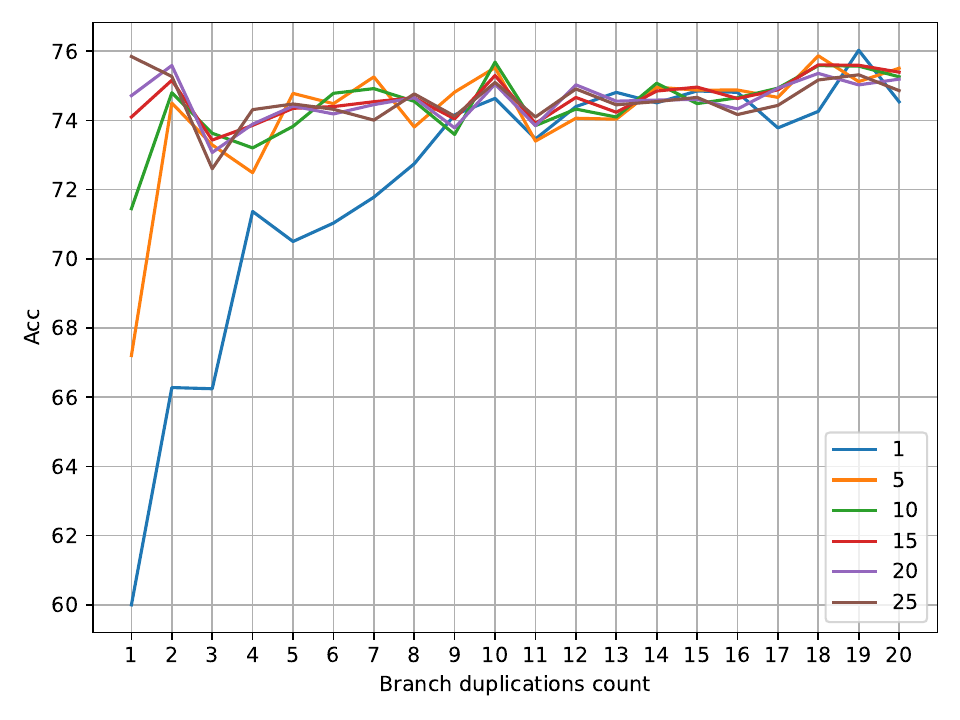}}
	\caption{Accuracy for each branch in MLA with  $ k $-invariant top-$ k $ cross-attention and repeated stack branches.}
	\label{fig:tk_branch}
\end{figure}

In the final component analysis, we explore the choice of ``MLA'' based on multiple branches. MLA indicates that each action is aligned to all of its descriptions at multiple levels. The weights of the projection at each level are not shared. Based on the observations made in the previous experiments, we used top-$ k $ attention to aggregate multiple descriptive semantic embeddings. The scoring network $ M $ changes from one to many, yielding a final performance of 80.96\% (Table \ref{tab:ex_description} line 8).
Figure \ref{fig:top_k} compares the number of branches and the corresponding top- $ k $ of MLA. The best performance has 12 branches, each equipped with a different $ k $, namely \{top-1, top-5, $  \cdots $ , top-60\}.
In addition, we give the accuracy results for repeated superposition of single branches with fixed $ k $ in Figure \ref{fig:tk_branch}. It is worth noting that this result is not equipped with $ A_{inv}$.
It can be seen that when $ k $ is constant, the model converges faster with increasing branches and slight fluctuation in performance.
When equipped with semantic features of different granularity, the fitting ability of the model improves significantly as the number of branches increases compared to the use of fixed granularity in Figure \ref{fig:tk_branch}.
This result suggests that increasing the discriminability and diversity of action representations impacts the ZSSAR task positively.

\subsection{Generalization Study}
In order to investigate the generalization ability of the proposed method, we conducted extended experiments on MLA, i.e., we added MLA components to other ZSSAR methods, and the results are shown in Table \ref{tab:general_mla}. Since the baseline methods DeViSE, RelationNet, and ReViSE do not distinguish between the label names' lexicality; we chose these methods for our experiments. The results show that MLA performs well based on the other frameworks. The experimental results on NTU-60 improved to 2.07\%, 3.88\%, and 2.94\%, respectively. 

In addition, we have selected a better-performing ZSL method for image classification, CuPL \cite{Pratt_2023_ICCV_CuPL}, as our baseline method for experiments on ImageNet. Since CuPL is a training-free method, we tested the effectiveness of our SFE module for synthetic semantic embedding on CuPL. Specifically, instead of averaging multiple prompt features per class as CuPL extracts them, we retain each class's prompt features. In inference, let the input samples compute the cosine similarity with all the prompt features, and then use these similarities as attention for prompt feature aggregation according to different top-k attention in MLA, and compute the cosine similarity between the image features and these aggregated features as the branching of the prediction scores. Finally, these scores are summed up as the prediction score for this class. After calculating all the category scores, the category with the highest score is the final prediction. After applying our method (with branch parameters \{4, 16, 32, 42, 46\}), the top1 accuracy of CuPL on ImageNet's validation set is improved from 76.69\% to 76.88 \%, fully validating our method's generalization.

\begin{figure}[t]
	\centering
	{\includegraphics[width=0.9\linewidth]{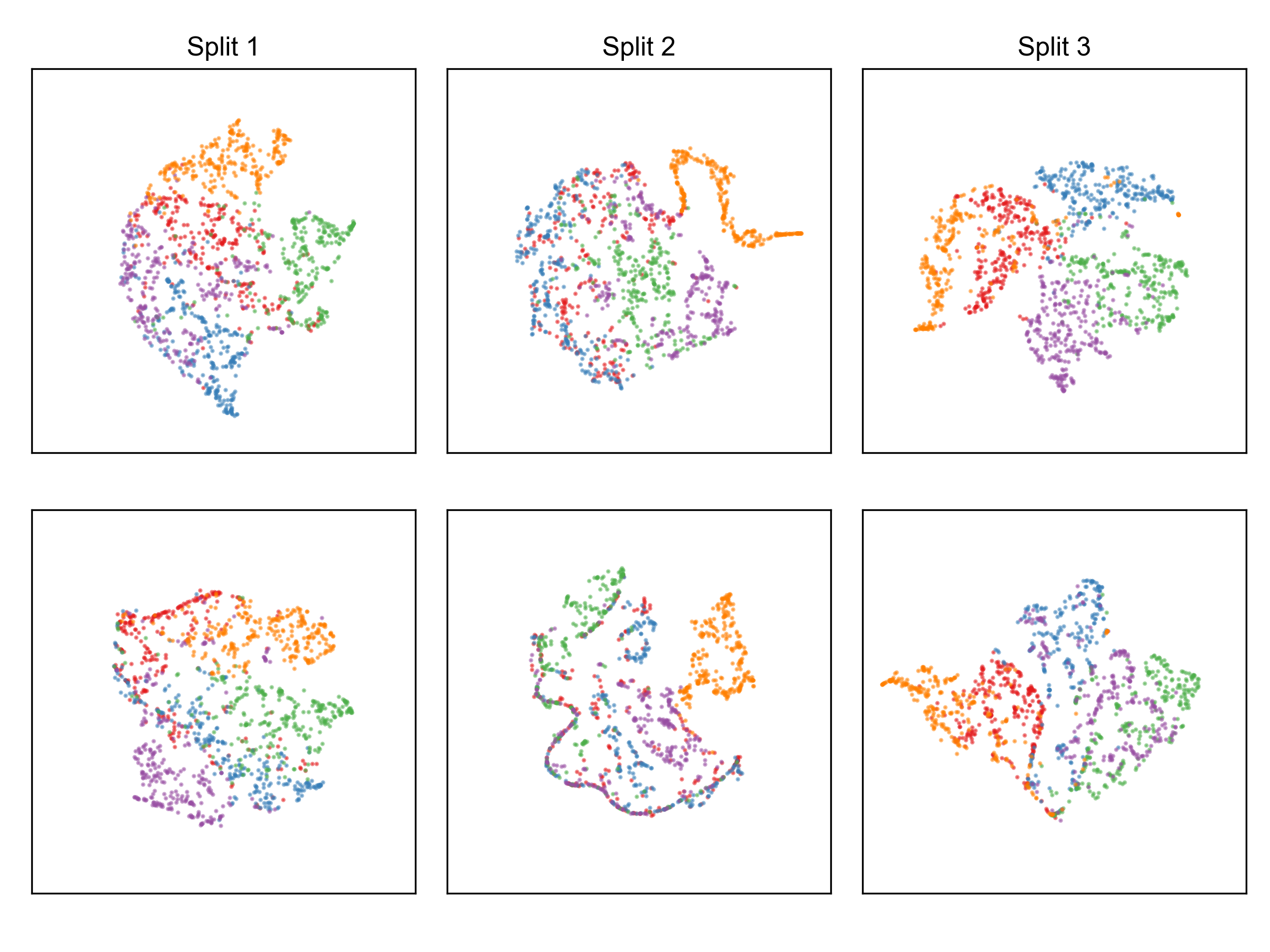}}
	\caption{The t-SNE visualization on NTU-60. The first line is the method SMIE, and the second is our method.}
	\label{fig:tsne}
\end{figure}

\subsubsection{Qualitative Results and Analysis}
\begin{figure}[t]
	\centering
	{\includegraphics[width=0.9\linewidth]{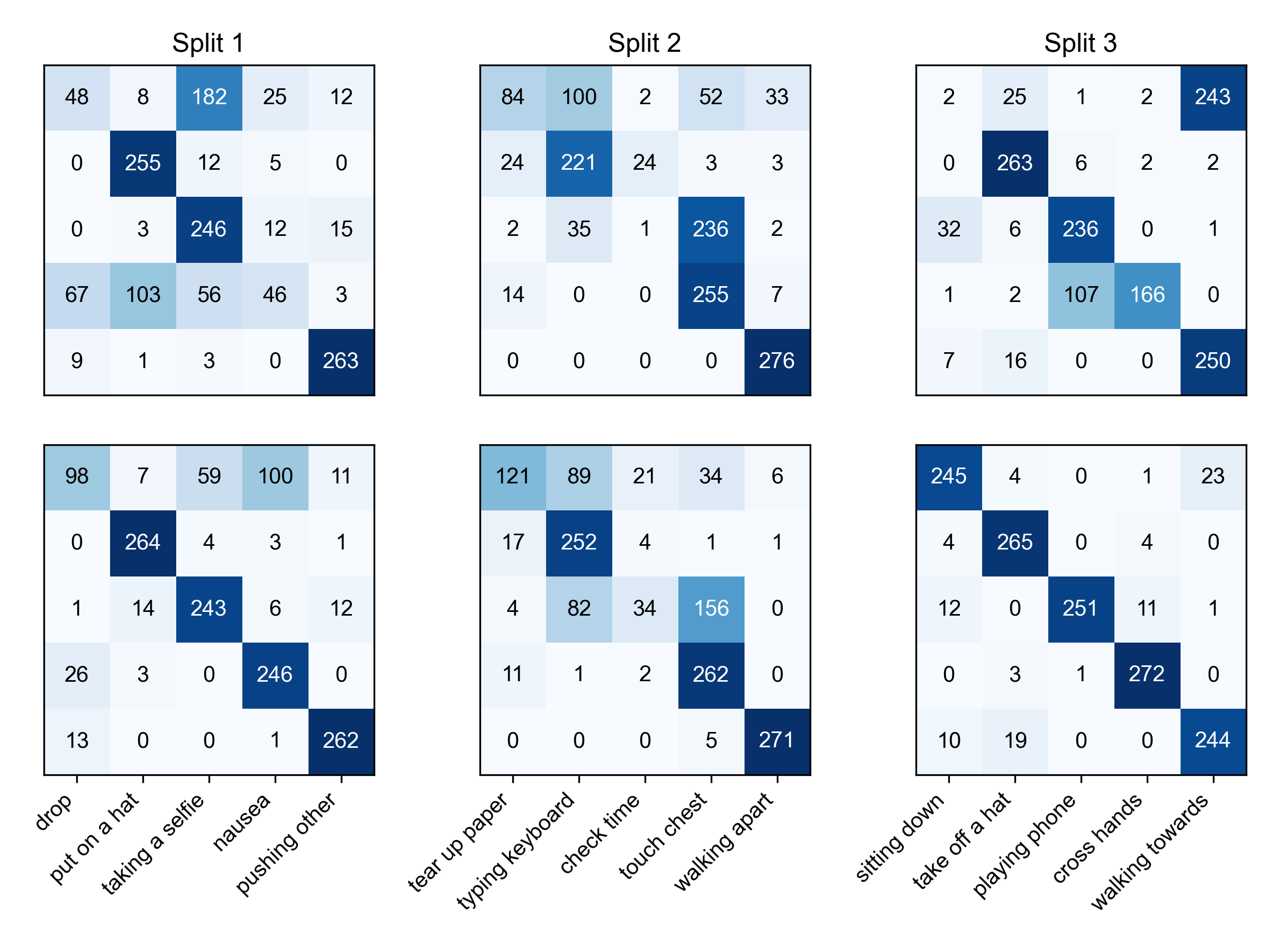}}
	\caption{The confusion matrix visualization on NTU-60. The first line is the method SMIE, and the second is our method.}
	\label{fig:visualization}
\end{figure}

We also use t-SNE \cite{Maaten2008VisualizingDU} to visualize the features learned by MLA. We plotted three sets of t-SNE graphs, one using labeled simple descriptions, one using the proposed global descriptions, and the other using multi-level augmented descriptions, to focus on the impact of fine-grained descriptions on feature learning. The t-SNE feature visualization in Fig. \ref{fig:tsne} shows that adding a reasonable global description improves visual and semantic alignment. In addition, we visualize the confusion matrix for the three splits on NTU-60 in Fig. \ref{fig:visualization}. The unseen classes form the matrices, and the value at the corresponding position is the number of samples classified into that class, thus allowing us to observe each category's classification accuracy. In these categories, many models of hand movements, such as ``tear up paper,'' ``typing keyboard,'' and ``check time'' are more prone to misclassification. Therefore, our subsequent work plans to align hand-specific embeddings.

\subsubsection{Computation Cost}
Compared with single-head scoring network SMIE, which is 575.5 MFLOPs, our multi-head InfoCPL is 591.2 MFLOPs. We used the popular python library thop to compute the FLOPs of the model. Although we have more branches, the central computation of the model is actually in the backbone part, so it is only 15.7 MFLOPs for multi-head $ M $. $ A_{inv} $ is the training-time strategy that does not increase FLOPs during inference.

\section{Conclusion}
\label{sec:conclusion}
This paper proposes the InfoCPL framework for zero-shot action recognition. In detail, we propose a SFE module for generating rich textual features beyond the previous work that only uses category descriptions. Consequently, we propose a multi-level alignment module to increase feature distinguishability. In addition, we further regularize the feature space by enriching the diversity of skeleton sequences through training and requiring different sequences belonging to a uniform class to be correctly aligned. The experiment results show that the proposed InfoCPL outperforms the current state-of-the-art methods, validating the design direction of ZSL for enriched class distinguishability information.

\bibliographystyle{IEEEtran}
\bibliography{example_paper}

%
%
%
%
%

\vfill

\end{document}